# Query-based versus Resource-based Cache Strategies in Tag-Based Browsing Systems

Joaquín Gayoso-Cabada, Mercedes Gómez-Albarrán, José-Luis Sierra

Universidad Complutense de Madrid, 28040 Madrid, Spain
{jgayoso, mgomeza, jlsierra}@ucm.es

**Abstract.** Tag-based browsing is a popular interaction model for navigating digital libraries. According to this model, users select descriptive tags to filter resources in the collections. Typical implementations of the model are based on inverted indexes. However, these implementations can require a considerable amount of set operations to update the browsing state. To palliate this inconvenience, it is possible to adopt suitable cache strategies. In this paper we describe and compare two of these strategies: (i) a *query-based* strategy, according to which previously computed browsing states are indexed by sets of selected tags; and (ii) a *resource-based* strategy, according to which browsing states are indexed by sets of filtered resources. Our comparison focused on runtime performance, and was carried out empirically, using a real-world web-based collection in the field of digital humanities. The results obtained show that the resource-based strategy clearly outperforms the query-based one.

## 1 Introduction

*Tag-based browsing* [7][27] is a popular interaction model adopted in many digital libraries [21]. According to this model, users can filter resources by employing descriptive *tags*. For this purpose, they can add new tags to shrink the current set of resources selected or to exclude tags in order to widen it. In consequence, the system updates the browsing state to provide: (i) the new set of filtered resources; and (ii) the new set of selectable tags.

A typical way to implement tag-based browsing is by using an *inverted index* [30]. An inverted index provides, for each tag, all the resources tagged with such a tag. In this way, after each user action the system can compute the new browsing state by performing several set operations involving the inverted index, the current browsing state, and, eventually, the overall underlying collection. Although there has been extensive research in performing these operations efficiently [5], the number of operations required can be appreciable, which can negatively impact user experience.

In order to decrease the cost of updating the browsing states, it is possible to use different cache strategies [8]. In particular, in this paper we will consider two of these strategies:

- *Query-based* strategy. This strategy uses the set of tags selected by the user (the *active tags*) as a cache index. Thus, it resembles the query-based caching mechanisms usually implemented in database systems [23]. The strategy is useful to identify *equivalent* browsing paths: two browsing paths are equivalent if they comprise the same set of active tags (although these tags may have been selected in a different order in each path). Since equivalent browsing paths lead to the same browsing state, by identifying a path to be equivalent to a previously explored one it is possible to reuse the cached information instead of re-computing it.
- *Resource-based* strategy. This strategy uses the set of filtered resources as the cache index. Thus, this strategy is able to detect paths leading to the same browsing state, even when they differ in their active tags. Therefore, the equivalency detection capability of this strategy outperforms that of the *query-based* one, since it is possible to have many distinct sets of active tags filtering the same set of resources. The disadvantage is the need to compute the set of filtered resources even for equivalent browsing paths.

In addition to describing these strategies, in this paper we will compare them using a real-world digital collection in the field of digital humanities (in particular, in the archeological domain). As it was mentioned above, and as we realized during our experience with several real-world collections in the field of digital humanities (see, for instance, [11]), response times during the updating of browsing states is a critical aspect that can directly impact user experience and satisfaction. Therefore, comparison will be focused on runtime performance, and, in particular, on the impact of each cache strategy on the time spent updating the information of the browsing states after each browsing action.

The rest of the paper is organized as follows: section 2 surveys some works related to ours. Section 3 describes tag-based browsing in more detail. Section 4 describes the two cache strategies. Section 5 describes the empirical evaluation results. Finally, section 6 provides some conclusions and lines of future work.

## 2 Related Work

Tag-based browsing naturally arises in digital collections organized as *semantic file systems* [28], in which resources are described using tags instead of being placed in particular folders or directories. In consequence, tag-based browsing resembles conventional directory-based navigation. Some works adopting this organization are [1][7][12][24][26]. These works typically use inverted indexes to organize the information and to speed up navigation. However, none of them discuss specific techniques concerning cache management in order to further enhance the browsing process.

Another field where tag-based browsing is extensively used is in digital collections supporting social tagging, in the style of Web 2.0 systems [6]. Some examples of works following this navigation model are [14][15][16][17][18][19][20]. Again, most of the systems described in this work use inverted indexes, but none of them focus on concrete techniques for dealing with the navigation cache.

In [8] we described an approach to enhancing tag-based browsing systems with inverted indexes and a cache strategy. By combining the cache-based and the resource-based strategy it is possible to obtain an improved version of the strategy proposed in that work. Finally, the work described in this paper is closely related to our previous work on *navigation automata* [9][10][11] for speeding up tag-based browsing. States in these automata correspond to sets of resources selected, while transitions correspond to the tags added by the users to shrink these sets. Navigation automata themselves are also closely related to *concept lattices* (as they are understood in *formal concept analysis* [3]), artifacts that have also been used to organize digital collections (e.g., [13][29]). The approaches described in this paper can be understood as a dynamic expansion of these automata and lattices, which amortizes the overhead of their explicit construction during browsing, while also avoiding the construction of parts that will never be explored by the user.

## 3    Tag-Based Browsing

In this section we develop the aspects concerning tag-based browsing in more detail. Subsection 3.1 details the tag-based digital collection model proposed. Subsection 3.2 details the interaction technique itself.

### 3.1    Tag-based Digital Collections

In this paper we will adopt a simple model of digital collection according to which a collection consists of:

| Resources | Annotation | Resources | Annotation |
|---|---|---|---|
| **r1** 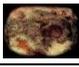 | `Cave-Painting`<br>`Cantabrian`<br>`Prehistoric` | **r4** 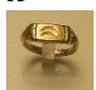 | `Tartesian`<br>`Plateau`<br>`Protohistoric` |
| **r2** 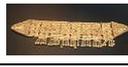 | `Cave-Painting`<br>`Levant`<br>`Prehistoric` | **r5** 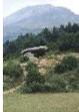 | `Phoenician`<br>`Penibaetic`<br>`Protohistoric` |
| **r3** 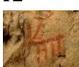 | `Megalithic`<br>`Cantabrian`<br>`Prehistoric` | **r6** 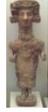 | `Punic`<br>`Levant`<br>`Protohistoric` |

**Fig. 1.** A small digital collection concerning prehistoric and protohistoric art in Spain

— A set of *digital resources*. Resources are digital objects whose nature is no longer constrained by the model (e.g., media files like images, sound, video, etc., external resources identified by their URIs, or entities of a more abstract nature, such as tu-ples of a table in a relational database, records in a bibliographical catalog, elements in an XML document, rows from a spreadsheet, etc.).
— A set of *descriptive tags*. These tags are used to describe the resources. Notice that this cataloguing model, although simple, is not excessively limiting, since it is always possible to think of tags as terms or concepts taken from more sophisticated cataloguing schemata. For instance, we have followed this approach in *Clavy*, an experimental platform for managing digital collections with *reconfigurable* cataloguing schemata [9][10][11] presented as reconfigurable hierarchies of *element* types. Since the hierarchical structure of *Clavy* schemata can change unexpectedly, the internal implementation of the *Clavy* browsing system was assimilated into a tag-based one (in *Clavy*, *tags* corresponded with *element-value* pairs).
— The *annotation* of the resources. This annotation consists of associating tags with resources, which effectively catalogues these resources and, therefore, enables future uses of the collection (navigation, search, etc.).

Fig. 1 outlines a small collection that follows this model. In this collection, resources are six image archives corresponding to artistic objects from the Prehistoric and Protohistoric artistic periods in Spain. Annotations are shown next to each resource. Tags describe the historic period (`Prehistoric` or `Protohistoric`), the artistic style (`Cave-Painting`, `Megalithic`, `Tartesian`, `Punic`, `Phoenician`), and the geographical area in which the object was discovered (`Cantabrian`, `Levant`, `Plateau` or `Penibaetic`).

### 3.2 The browsing model

As indicated earlier, the tag-based browsing model adopted in this paper will allow the user to focus on a set of resources by adding and excluding descriptive tags. For this purpose, the user can carry out two different kind of actions:

| Operation | Intended meaning |
|---|---|
| `resources()` | It provides all the resources in the collection. |
| `tags()` | It provides all the tags in the collection. |
| `selectable_tags(R,T)` | It determines which tags in **T** are selectable tags for the set of resources **R** (i.e., tags annotating *some*, but *not all*, of the resources in **R**). |
| `next_user_action()` | It returns the next interaction action carried out by the user (it will return ⊥ when the user finishes interaction). |
| `filter(R,t)` | It returns all those resources in **R** annotated with *t* |
| `query(F)` | It returns the resources in the collection annotated with all the tags in **F** |

**Fig. 2.** Basic operations supporting tag-based browsing

— Adding a tag *t* to the set of *active tags*. This *add* action will be denoted by +*t*.
— Removing the tag *t* from the set of active tags. This *remove* action will be denoted by ×*t*.

The browsing system will maintain a *browsing state* uniquely determined by the set of active tags **F**. This state will have the following information items associated:

- The set of resources $\mathbf{R^F}$ filtered by $\mathbf{F}$ (i.e., those resources in the collection annotated with all the tags in $\mathbf{F}$).
- The set of selectable tags $\mathbf{S^F}$ that can intervene in an *add* action (i.e., those tags the user can use to further winnow down the current set of filtered resources). Each tag in $\mathbf{S^F}$ will annotate some (but not all) the resources in $\mathbf{R^F}$.

```
F ← ∅
R^F ← resources()
S^F ← selectable_tags(R^F,tags())
do
   user_action ← next_user_action()
   if user_action = +t
      F ← F ∪ {t}
      R^F ← filter(R^F,t)
      S^F ← selectable_tags(R^F,S^F-{t})
   else if user_action = ×t
      F ← F - {t}
      R^F ← query(F)
      S^F ← selectable_tags(R^F, tags()-F)
   end if
until user_action = ⊥
```

**Fig. 3.** Algorithmic description of the tag-based browsing behavior

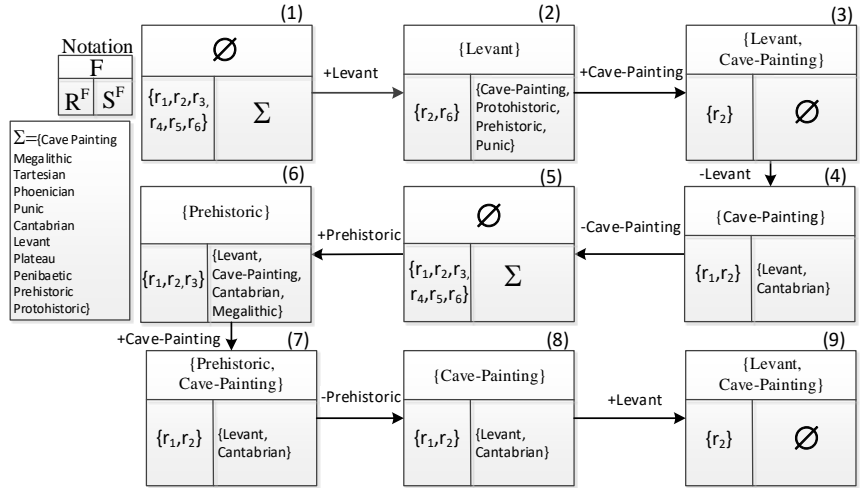

**Fig. 4.** A browsing session for the collection in Fig. 1

Fig. 2 introduces a set of primitive operations in terms of which the browsing behavior can be described, and outlines its intended meaning. Fig. 3 describes the browsing behavior itself. In this way:

- The browsing process begins by setting $\mathbf{R}^F$ to all the resources in the collection, and $\mathbf{S}^F$ to all the tags able to shrink these resources. Subsequently, the process proceeds until there are no more user actions available.
- To execute a +*t* action, it is enough to: (i) add *t* to the set $\mathbf{F}$ of active tags; (ii) filter the resources in $\mathbf{R}^F$ to those tagged with *t*; and (iii) update $\mathbf{S}^F$ to the corresponding selectable tags (notice that *t* can be excluded as a selectable tag, since it will be shared by all the resources in the updated $\mathbf{R}^F$; in addition, notice that all the resulting selectable tags must already be selectable *before* updating $\mathbf{S}^F$).
- To execute a ×*t* action, the steps to be carried out are: (i) remove *t* from $\mathbf{F}$; (ii) set $\mathbf{R}^F$ to all those resources tagged by all the tags in the updated $\mathbf{F}$; and (iii) update $\mathbf{S}^F$ (in this case, the potentially selectable tags are all the tags in the collection, with the exception of those in the updated $\mathbf{F}$).

Fig. 4 outlines a tag-based browsing session for the collection in Fig. 1.

## 4      Cache Strategies

Updating the $\mathbf{R}^F$ and, especially, the $\mathbf{S}^F$ sets can be a costly process, since it can require several set operations (this fact is made apparent, for instance, in [8], where this process is implemented using inverted indexes). Therefore, the adoption of cache strategies like those aforementioned can substantially speed up this process. In this section we detail the abovementioned *query-based* strategy (subsection 4.1) and the *resource-based* one (subsection 4.2).

### 4.1     Query-based Strategy

As suggested in section 1, the query-based strategy binds sets of active tags $\mathbf{F}$ to the information for the resulting browsing state (i.e., the $\mathbf{R}^F$ and the $\mathbf{S}^F$ sets). Fig. 5 describes the basic operations for managing the query-indexed cache, i.e., to store and retrieve the information. Using these operations, Fig. 6 describes the browsing strategy by extending the basic tag-based browsing process in Fig. 2 with cache capabilities. For this purpose, when the browsing system executes a browsing action:

| Operation | Intended meaning |
|---|---|
| `cache_by_query(F,R`$^\mathbf{F}$`,S`$^\mathbf{F}$`)` | It stores $\mathbf{R}^F$ and $\mathbf{S}^F$ in the cache. |
| `retrieve_by_query(F)` | It retrieves the ($\mathbf{R}^F$, $\mathbf{S}^F$) pair from the cache. If $\mathbf{F}$ is not cached, it returns ⊥ |

**Fig. 5.** Operations for managing the query-indexed cache

- Firstly, it updates $\mathbf{F}$ accordingly (adding *t* for +*t* actions, and removing *t* for ×*t* actions).
- Afterwards, it uses the updated $\mathbf{F}$ to query the cache in the hope of retrieving the $\mathbf{R}^F$ and $\mathbf{S}^F$ values. If it fails to retrieve these values, (i) it updates them like the basic, un-cached, process described in Fig. 3; and (ii) it caches the updated values. Otherwise, it uses the retrieved values to update $\mathbf{R}^F$ and $\mathbf{S}^F$.

```
F ← ∅
R^F ← resources()
S^F ← selectable_tags(R^F,tags())
cache_by_query(F, R^F, S^F)
do
   user_action ← next_user_action()
   if user_action ≠ ⊥
      if user_action = +t
         F ← F ∪ {t}
      else, let user_action = ×t in
         F ← F - {t}
      end if
      cached_info ← retrieve_by_query(F)
      if cached_info = ⊥
         if user_action = +t
            R^F ← filter(R^F,t)
            S^F ← selectable_tags(R^F, S^F-{t})
         else, let user_action = ×t in
            R^F ← query(F)
            S^F ← selectable_tags(R^F, tags()-F)
         end if
         cache_by_query(F, R^F, S^F)
      else
         (R^F, S^F) ← cached_info
      end if
   end if
until user_action = ⊥
```

**Fig. 6.** Tag-based browsing with a query-indexed cache

Therefore, each time the **F** set is updated, previously to computing the $\mathbf{R}^F$ and $\mathbf{S}^F$ sets, the cache is consulted. If this computation is finally carried out, this information is cached for subsequent use. In addition, since the cache indexes are the **F** sets of active tags, the order in which the tags are selected does not matter.

The strategy makes it possible, for instance, to avoid the computation of $\mathbf{R}^F$ and $\mathbf{S}^F$ in state (9) of Fig. 4, regardless of the fact that, when this information was cached -state (3)-, the sequence of tags selected was Levant followed by Cave-Painting, while in state (9) the selection order was inverted: first Cave-Painting, then Levant. It illustrates how this strategy is able to successfully deal with equivalent browsing paths, in the sense of leading to identical **F** sets. Concerning the other browsing states in Fig. 4, the strategy also avoids re-computing the $\mathbf{R}^F$ and $\mathbf{S}^F$ sets in states (5) and (8).

### 4.2 Resource-based Strategy

While the query-based strategy successfully deals with browsing paths leading to identical sets of active tags, it fails to detect where two distinct sets of active tags filter the same set of resources. For instance, in the collection of Fig. 1 {Cave-Painting} and {Prehistoric, Cave-Painting} filter the same set of resources ({**r1**, **r2**}). However, the query-based strategy will be unaware of this fact. For instance, in state (7) of Fig. 4 it will re-compute $\mathbf{R}^F$ and $\mathbf{S}^F$, although these sets were already computed for state (4). The resource-based strategy alleviates this shortcoming. Fig. 7 describes the resource-indexed cache managing operations required to implement the strategy. Fig. 8 describes the resulting algorithmic behavior. In this case, the execution of a browsing action involves:

| Operation | Intended meaning |
| --- | --- |
| `cache_by_resources(R,S)` | It binds the set of selectable tags **S** to the set of filtered resources **R** in the cache. |
| `retrieve_by_resources(R)` | It retrieves the set of selectable tags **S** for the resources **R** from the cache. If **R** is not cached, it returns ⊥ |

**Fig. 7.** Operations for managing the resources-indexed cache

```
F ← ∅
R^F ← resources()
S^F ← selectable_tags(R^F, tags())
cache_by_resources(R^F, S^F)
do
   user_action ← next_user_action()
   if user_action ≠ ⊥
      if user_action = +t
         F ← F ∪ {t}
         R^F ← filter(R^F, t)
      else, let user_action = ×t in
         F ← F - {t}
         R^F ← query(F)
      end if
      cached_sel_tags ← retrieve_by_resources(R^F)
      if cached_sel_tags = ⊥
         if user_action = +t
            S^F ← selectable_tags(R^F, S^F-{t})
         else, let user_action = ×t in
            S^F ← selectable_tags(R^F, tags()-F)
         end if
         cache_by_resources(R^F, S^F)
      else
         S^F ← cached_sel_tags
      endif
   end if
until user_action = ⊥
```

**Fig. 8.** Tag-based browsing with a resource-indexed cache

— Firstly, updating **F** as in the query-based strategy in Fig. 6.
— Secondly, computing $\mathbf{R}^F$ as in the basic, un-cached, strategy in Fig. 3.
— Finally, querying the cache with $\mathbf{R}^F$ in the hope of getting a value for $\mathbf{S}^F$. If the query fails, $\mathbf{S}^F$ is computed as in Fig. 3, and then cached. Otherwise, $\mathbf{S}^F$ is updated to the value retrieved.

Therefore, notice that the resource-based strategy only avoids re-computing the $\mathbf{S}^F$ set, since it uses the set of filtered resources $\mathbf{R}^F$ as the cache index. However, as described in [8], the computation of $\mathbf{R}^F$ is usually much more agile than that of $\mathbf{S}^F$. Indeed:

— Once $\mathbf{R}^F$ is updated, the computation of $\mathbf{S}^F$ requires simulating the effect of an *add* action for *each* potentially selectable tag, i.e. (i) for each tag in the previous value for $\mathbf{S}^F$ ($t$ excluded) if the user action was +$t$; or (ii) for each *non-active* tag (i.e., for each tag in the collection not present in $\mathbf{F}$) if the user action was a *remove* action.
— In consequence, the complexity of updating $\mathbf{S}^F$ is typically an order of magnitude greater than updating $\mathbf{R}^F$ (by using an inverted index, $\mathbf{R}^F$ can be updated by: (i) intersecting it with the entry in the inverted index for $t$ if the user action was +$t$; or (ii) intersecting the entries for each surviving active tag if the action was ×$t$).

Concerning the browsing session in Fig. 4, the resource-based strategy will avoid re-computing $\mathbf{S}^F$ in (7), as well as in (5), (8) and (9).

## 5     Evaluation Results

This section describes an empirical comparison between the query-based and the resource-based cache strategies oriented to assessing the differences in runtime performance between the two strategies. Subsection 5.1 describes the experimental setting. Subsection 5.2 presents the comparison results.

### 5.1    Experimental Setting

In order to carry out the comparison we implemented the two cache strategies in the aforementioned *Clavy* platform. In this context, we set an experiment concerning *Chasqui* [25],[1] a digital collection on Pre-Columbian American archeology. The current *Chasqui* version in *Clavy* consists of 2060 resources. The experimental setting allowed us to randomly generate browsing sessions. The generation process was based on the following random browsing process:

— In each browsing state in which both *add* and *remove* actions were allowed, a first, random decision on whether to choose one or another type of action was made (equal probability of 0.5 for each choice).
— The generation of *add* actions was carried out by performing a second random decision, oriented to deciding whether to pick a tag between the 20 first selectable ones (probability of 0.8), or whether to pick a less significant tag (those placed from position 21 onwards). Finally, once the segment was determined, a tag in this segment was picked with equal probability.

---

[1] http://oda-fec.org/ucm-chasqui

– Concerning *remove* actions, random selection prioritized the most recently added tags: the last *k*-th added tag was picked with probability $0.8^{k-1} \times 0.2$ (i.e., following a geometric distribution with success probability p=0.2).

These randomly generated browsing sessions allowed us to empirically check whether there were differences in the performances of the two strategies. Since our study was focused on runtime performance, we choose the total time spent for updating the browsing states during each browsing session (*cumulative browsing time*) as the primary comparison metric, since it provides all the time spent updating the browsing snapshot during each browsing session in a single measure. On the other hand, it is important to highlight that we were trying to assess which strategy provides responses faster than which others, not whether a particular tag-based collection (*Chasqui*, in this case) was correctly catalogued. This focus on time efficiency left out other resource and quality-based metrics which, like *precision* and *recall*, are typically used to evaluate information retrieval systems [22]. Indeed, both approaches (query-based and resource-based) are observationally equivalent from the point of view of the tag-based browsing model (i.e., when applied on the same sequence of browsing actions, both approaches will produce exactly the same sequence of browsing states, with the same $\mathbf{R}^F$ and $\mathbf{S}^F$ sets), which made it pointless to apply IR metrics like *recall* and *precision* in this study.

Next subsection summarizes the comparison results[2].

## 5.2 Comparison results

The execution traces of the two strategies clearly suggested that the resource-based strategy outperformed the query-based one. For instance, Fig. 9 shows a representative example of these traces, displaying the cumulative browsing times for both strategies and for a browsing session consisting of 10000 user actions. In this particular execution the total computation time spent by the resource-based strategy (about 0.68 seconds) decreased the total time of the query-based approach (about 1.53 seconds) by about 55%.

After successive repetitions of the experiment, we observed similar behaviors, with improvements in favor of the resource-based strategy. More concretely, we measured the total execution times of both strategies for 500 browsing sessions, each one also containing 10000 user actions. The resource-based approach lowered the execution time in a range between 42% and 69% (Fig. 10). The average enhancement was of 54.8% (95% CI [54,8, 55,02]%). After running a Wilcoxon signed-rank test to compare the cumulative browsing times for the two strategies (query-based, mean rank=.00, and resource-based, mean rank =250.5), we realized that this perceived difference was clearly statistically significant (Z = -19.375, p= .000).

---

[2] All the measures reported were taken on a machine with an Intel® Core™ i5-4660S 2.9GHz processor, RAM 16 GB and Windows 10 OS. Browsing software was programmed in Java. The browsing cache was maintained in memory using Java's `HashMaps`. Sets were managed using *roaring bitmaps* [4].

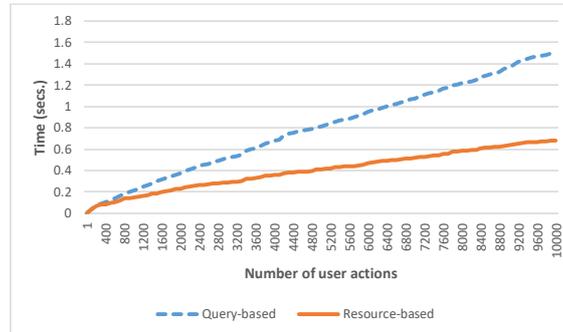

**Fig. 9.** Cumulative times for the query-based and the resource-based strategies for a randomly-generated browsing session

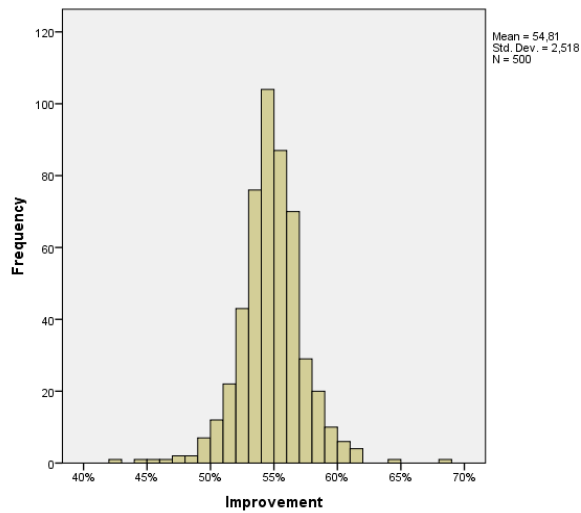

**Fig. 10.** Histogram of the improvement in time of the resource-based strategy

## 6      Conclusions and Future Work

In this paper we have described and compared two basic techniques to organize the browsing cache in tag-based digital collections. The first one, query-based strategy, is focused on caching browsing paths as set of active tags. In consequence, it is suitable for detecting equivalent browsing paths that involve exactly the same tags. However, it fails to detect paths that differ in some tag but that lead to the same set of resources. This shortcoming is overcome by the resource-based strategy, by using sets of filtered resources, instead of sets of active tags, as indexing items. As side effects, it always re-computes the set of filtered resources. The experimental results obtained in this paper make it apparent that the resource-based approach can actually outperform the query-based one. Nevertheless, it does not exempt us from accepting this result with caution. In particular, the result is heavily dependent on the particular tag-based browsing model adopted. For instance, we can think of more restricted tag-based browsing models, like those followed in semantic file systems. Indeed, since in a semantic file system, browsing resembles the typical operations in a conventional file system, active tags can only be removed in reverse order to how they are added. Notice that, under this assumption, the query-based approach could provide better performance, since after *remove* actions, the information for the resulting state will always be in the query-indexed cache.

Currently we are working on improving the treatment of *remove* actions in order to reduce the number of set operations required. We are also working on combining the caching strategies with our previous work in navigation automata, as well as generalizing the approach to enable the navigation through links among resources. Also, we plan to combine browsing and Boolean searches, allowing users to browse search results according to the tag-based browsing model. Finally, we also plan to carry out a more exhaustive evaluation. On the one hand, our aim is to improve the external validity of the results. For this purpose, we plan to replicate the experiment with additional collections in the Digital Humanities field (in particular, the collections referred to in [11]). We have already run the browsing session generation and simulation processes on those collections, and we have observed significant differences in the performance of the resource-based strategy with respect to the query-based one. While it provides evidence to the external validity of the results, at least in reference to the domain of digital humanities, data must be formally analyzed, and the differences observed formally tested. In addition, we plan to deal with collections outside the Digital Humanities domain (in particular, medical collections of clinical cases [2]) with the aim of widening the external validity of the results to other knowledge fields. Finally, it would also be nice to consider other models for tag-based browsing (e.g., the aforementioned models of semantic file systems). On the other hand, we also want to improve the ecological validity of our study. While the use of real-world collections gave us positive evidence in favor of ecological validity, the use of simulated browsing traces could negatively affect such validity dimension. Therefore, we plan to carry out the simulation of more realistic user browsing behavior models, as well as real-user browsing traces. In addition, it would also be interesting to use additional metrics (e.g., the maximum response time for all that registered in a browsing trace), as well as experiments focusing on measuring user satisfaction.

## Acknowledgements

This research is supported by the research projects TIN2014-52010-R and TIN2017-88092-R. Also, we would like to thank Mercedes Guinea and Alfredo Fernández-Valmayor (El Caño Foundation, Panamá), for their work on *Chasqui*.

# References

1. Bloehdorn, S.; Görlitz, O.; Schenk, S.; Völkel, M. TagFS - Tag Semantics for Hierarchical File Systems. In Proceedings of the 6th International Conference on Knowledge Management (I-KNOW 06). 2006
2. Buendía, F.; Gayoso-Cabada, J.; Sierra, J-L. Using Digital Medical Collections to Support Radiology Training in E-learning Platforms. In Proceedings of the 13th European Conference on Technology-Enhanced Learning (EC-TEL'18). 2018
3. Carpineto, C., Romano, G. Concept Data Analysis: Theory and Applications. Wiley. 2004
4. Chambi, S.; Lemire, D.; Kaser, O.; Godin, R. Better bitmap performance with Roaring bitmaps. Software-Practice&Experience, 46(5), 709-719. 2016
5. Culpepper, J-S.; Moffat, A. Efficient Set Intersection for Inverted Indexing. ACM Transations on Information Systems 29(1), article 1. 2010
6. Dimitrov, D.; Helic, D.; Strohmaier, M. Tag-Based Navigation and Visualization. In Brusilovsky, P.; He, D. (Eds.). Social Information Access, LNCS 10100, pp. 181–212. Springer 2018.
7. Eck, O.; Schaefer, D. A semantic file system for integrated product data management. Advanced Engineering Informatics, 25(2), 177-184. 2011
8. Gayoso-Cabada, J.; Gómez-Albarrán, M.; Sierra, J-L. Tag-Based Browsing of Digital Collections with Inverted Indexes and Browsing Cache. In Proceedings of the 6th Edition of the Technological Ecosystems for Enhancing Multiculturality Conference (TEEM'18). 2018
9. Gayoso-Cabada, J.; Rodríguez-Cerezo, D.; Sierra, J.-L. Multilevel Browsing of Folksonomy-Based Digital Collections. In Proceedings of the 17th Conference on Web Information Systems Engineering (WISE'16). 2016
10. Gayoso-Cabada, J.; Rodríguez-Cerezo, D.; Sierra, J-L. Browsing Digital Collections with Reconfigurable Faceted Thesauri. 25th International Conference on Information Systems Development (ISD). Katowize, Poland. 2016
11. Gayoso-Cabada, J.; Rodríguez-Cerezo, D.; Sierra, J-L. Browsing Digital Collections with Reconfigurable Faceted Thesauri. In Goluchowski, J. et al (Eds.) Complexity in Information System Development (extended and revised ISD'18 papers), 69-86. Lecture Notes in Information Systems and Organization. Springer. 2017
12. Gifford, D.K.; Jouvelot, P.; Sheldon, M.A.; O'Toole, J.W. Semantic file systems. SIGOPS Operating Systems Review, 25(5), 16-25. 1991
13. Greene, G-J., Dunaiski, M., Fischer, B. Browsing Publication Data using Tag Clouds over Concept Lattices Constructed by Key-Phrase Extraction. In Proceedings of Russian and South African Workshop on Knowledge Discovery Techniques Based on Formal Concept Analysis (RuZA'15). 2015.
14. Helic, D.; Trattner, C.; Strohmaier, M.; Andrews, K. On the navigability of social tagging systems. In: 2010 IEEE Second International Conference on Social Computing (SocialCom'10), pp. 161–168. 2010
15. Hernandez, M-E.; Falconer, S-M.; Storey, M-A.; Carini, S.; Sim, I. Synchronized tag clouds for exploring semi-structured clinical trial data. In Proceedings of the 2008 conference of the center for advanced studies on collaborative research: meeting of minds (CASCON'08). 2008
16. Kammerer, Y.; Nairn, R.; Pirolli, P.; Chi, E.H. Signpost from the masses: learning effects in an exploratory social tag search browser. In Proceedings of the SIGCHI Conference on Human Factors in Computing Systems (CHI'09). 2009

17. Kleinberg, J. Navigation in a small world. Nature 406(6798), 845. 2000
18. Koutrika, G.; Zadeh, Z-M.; Garcia-Molina, H. CourseCloud: summarizing and refining keyword searches over structured data. Proc. of the 12$^{th}$ International Conference on Extending Database Technology (EDBT), pp. 1132-1135. 2009
19. Leone, S.; Geel, M.; Müller, C.; Norrie, M.C. Exploiting Tag Clouds for Database Browsing and Querying. In the CaiSE Forum 2010 (Selected Extended Papers), Lecture Notes in Business Information Processing, 72. 2011
20. Lin, Y-L., Brusilovsky, P.; He; D. Finding cultural heritage images through a Dual-Perspective Navigation Framework. Inf. Proc. and Management, 52(5), 820-839. 2016
21. Redden, C.S. Social Bookmarking in Academic Libraries: Trends and Applications. The Journal of Academic Librarianship, 36(3), 219-227. 2010
22. Salton, G.; McGill, M.J. Introduction to Modern Information Retrieval. McGraw-Hill. 1986
23. Schwartz B., Tkachenko, V., Zaitsev, P. High Performance MySQL 3rd Edition. O'Reilly Media. 2012
24. Seltzer, M., Murphy, N. Hierarchical file systems are dead. In Proceedings of the 12th conference on Hot topics in operating systems (HotOS'09). 2009
25. Sierra, J-L.; Fernández-Valmayor, A.; Guinea, M.; Hernanz, H. From Research Resources to Learning Objects: Process Model and Virtualization Experiences. Educational Technology & Society 9(3), 56-68. 2006
26. Sim, H.; Kim, Y.; Vazhkudai,S.S.; Vallée, G.R.; Lim, S-H.; Butt, A.R. Tagit: an integrated indexing and search service for file systems. In Proceedings of the International Conference for High Performance Computing, Networking, Storage and Analysis (SC '17). 2017
27. Trattner, C.; Lin, Y.; Parra, D.; Yue, Z.; Real, W.; Brusilovsky, P. 2012. Evaluating tag-based information access in image collections. In Proceedings of the 23rd ACM conference on Hypertext and social media (HT '12), 113-122. 2012
28. Watson, R.; Dekeyser, S.; Albadri, N. Exploring the design space of metadata-focused file management systems. In Proceedings of the Australasian Computer Science Week Multiconference (ACSW '17). 2017
29. Way, T., Eklund, P. Social Tagging for Digital Libraries using Formal Concept Analysis. In Proceedings of the 17th International Conference on Concept Lattices and their Applications (CLA'10). 2010
30. Zobel, J.; Moffat, A. Inverted Files for Text Search Engines. ACM Computing Surveys 33(2), article 6. 2006